\documentclass[runningheads]{llncs}
\usepackage{graphicx}
\usepackage{amsmath,amssymb} 
\usepackage{color}
\usepackage{multirow}
\usepackage{booktabs}
\usepackage{url}

\begin{document}
\title{Predicting Gaze in Egocentric Video by Learning Task-dependent Attention Transition} 

\titlerunning{Predicting Gaze in Egocentric Video}
\author{Yifei Huang\inst{1} \and
Minjie Cai\inst{2,1}\thanks{corresponding author} \and
Zhenqiang Li\inst{1} \and
Yoichi Sato\inst{1}}
%
\authorrunning{Yifei Huang, Minjie Cai, Zhenqiang Li and Yoichi Sato}
%

\institute{$^1$The University of Tokyo, Tokyo, Japan~~ $^2$Hunan University, Changsha, China\\
\email{\{hyf,cai-mj,lzq,ysato\}@iis.u-tokyo.ac.jp}}
\maketitle              
\begin{abstract}
We present a new computational model for gaze prediction in egocentric videos by exploring patterns in temporal shift of gaze fixations (attention transition) that are dependent on egocentric manipulation tasks.
Our assumption is that the high-level context of how a task is completed in a certain way has a strong influence on attention transition and should be modeled for gaze prediction in natural dynamic scenes.
Specifically, we propose a hybrid model based on deep neural networks which integrates task-dependent attention transition with bottom-up saliency prediction. 
In particular, the task-dependent attention transition is learned with a recurrent neural network to exploit the temporal context of gaze fixations, \textit{e.g.} looking at a cup after moving gaze away from a grasped bottle.
Experiments on public egocentric activity datasets show that our model significantly outperforms state-of-the-art gaze prediction methods and is able to learn meaningful transition of human attention.

\end{abstract}
\section{Introduction}

With the increasing popularity of wearable or action cameras in recording our life experience, egocentric vision \cite{betancourt2015evolution}, which aims at automatic analysis of videos captured from a first-person perspective \cite{kitani2011fast}\cite{cai2015scalable}\cite{cai2017ego}, has become an emerging field in computer vision. In particular, as the camera wearer's point-of-gaze in egocentric video contains important information about interacted objects and the camera wearer's intent \cite{huang2017temporal}, gaze prediction can be used to infer important regions in images and videos to reduce the amount of computation needed in learning and inference of various analysis tasks \cite{fathi2012learning}\cite{xu2015gaze}\cite{cai2016understanding}\cite{cai2018desktop}. 

This paper aims to develop a computational model for predicting the camera wearer's point-of-gaze from an egocentric video. Most previous methods have formulated gaze prediction as the problem of saliency detection, and computational models of visual saliency have been studied to the find image regions that are likely to attract human attention. The saliency-based paradigm is reasonable because it is known that highly salient regions are strongly correlated with actual gaze locations \cite{parkhurst2002modeling}. However, the saliency model-based gaze prediction becomes much more difficult in natural dynamic scenes, \textit{e.g.} cooking in a kitchen, where high-level knowledge of the task has a strong influence on human attention.

In a natural dynamic scene, a person perceives the surrounding environment with a series of gaze fixations which point to the objects/regions related to the person's interactions with the environment. It has been observed that the attention transition is deeply related to the task carried out by the person. Especially in object manipulation tasks, the high-level knowledge of an undergoing task determines a stream of objects or places to be attended successively and thus influences the transition of human attention. For example, to pour  water from a bottle to a cup, a person always first looks at the bottle before grasping it and then change the fixation onto the cup during the action of pouring. Therefore, we argue that it is necessary to explore the task-dependent patterns in attention transition in order to achieve accurate gaze prediction.

In this paper, we propose a hybrid gaze prediction model that combines bottom-up visual saliency with task-dependent attention transition learned from successively attended image regions in training data. The proposed model is mainly composed of three modules. The first module generates saliency maps directly from video frames. It is based on a two-stream Convolutional Neural Network (CNN) which is similar to traditional bottom-up saliency prediction models. The second module is based on a recurrent neural network and a fixation state predictor which generates an attention map for each frame based on previously fixated regions and head motion. It is built based on two assumptions. Firstly, a person's gaze tends to be located on the same object during each fixation, and a large gaze shift almost always occurs along with large head motion \cite{land2004coordination}. Secondly, patterns in the temporal shift between regions of attention are dependent on the performed task and can be learned from data. The last module is based on a fully convolutional network which fuses the saliency map and the attention map from the first two modules and generates a final gaze map, from which the final prediction of 2D gaze position is made.

Main contributions of this work are summarized as follows:
\begin{itemize}
    \item We propose a new hybrid model for gaze prediction that leverages both bottom-up visual saliency and task-dependent attention transition. 
    \item We propose a novel model for task-dependent attention transition that explores the patterns in the temporal shift of gaze fixations and can be used to predict the region of attention based on previous fixations.
    \item The proposed approach achieves state-of-the-art gaze prediction performance on public egocentric activity datasets.
\end{itemize}

\section{Related Works}

\subsubsection{Visual Saliency Prediction.} 
Visual saliency is a way to measure image regions that are likely to attract human attention and thus gaze fixation \cite{borji2013state}. Traditional saliency models are based on the feature integration theory \cite{treisman1980feature} telling that an image region with high saliency contains distinct visual features such as color, intensity and contrast compared to other regions. After Itti \emph{et al.}'s primary work \cite{itti1998model} on a computational saliency model, various bottom-up computational models of visual saliency have been proposed such as a graph-based model \cite{harel2007graph} and a spectral clustering-based model \cite{hou2012image}. Recent saliency models \cite{lin2014saliency}\cite{huang2015salicon}\cite{pan2016shallow} leveraged a deep Convolutional Neural Network (CNN) to improve their performance. 
More recently, high-level context has been considered in deep learning-based saliency models. In \cite{simonyan2013deep}\cite{cao2015look}, class labels were used to compute the partial derivatives of CNN response with respect to input image regions to obtain a class-specific saliency map. In \cite{zhao2015saliency}, a salient object is detected by combining global context of the whole image and local context of each image superpixel. In \cite{ramanishka2017top}, region-to-word mapping in a neural saliency model was learned by using image captions as high-level input. 

However, none of the previous methods explored the patterns in the transition of human attention inherent in a complex task. In this work, we propose to learn the task-dependent attention transition on how gaze shifts between different objects/regions to better model human attention in natural dynamic scenes. 

\subsubsection{Egocentric Gaze Prediction.} 
Egocentric vision is an emerging research domain in computer vision which focuses on automatic analysis of egocentric videos recorded with wearable cameras. Egocentric gaze is a key component in egocentric vision which benefits various egocentric applications such as action recognition \cite{fathi2012learning} and video summarization \cite{xu2015gaze}. 
Although there is correlation between visually salient image regions and gaze fixation locations \cite{parkhurst2002modeling}, it has been found that traditional bottom-up models for visual saliency is insufficient to model and predict human gaze in egocentric video \cite{yamada2010can}. 
Yamada \emph{et al.} \cite{yamada2011attention} presented a gaze prediction model by exploring the correlation between gaze and head motion. In their model, bottom-up saliency map is integrated with an attention map obtained based on camera rotation and translation to infer final egocentric gaze position.
Li \emph{et al.} \cite{li2013learning} explored different egocentric cues like global camera motion, hand motion and hand positions to model egocentric gaze in hand manipulation activities. They built a graphical model and further combined the dynamic behaviour of gaze as latent variables to improve the gaze prediction. However, their model is dependent on predefined egocentric cues and may not generalize well to other activities where hands are not always involved.
Recently, Zhang \emph{et al.} \cite{zhang2017deep} proposed the gaze anticipation problem in egocentric videos. In their work, a Generative Adversarial Network (GAN) based model is proposed to generate future frames from a current video frame, and gaze positions are predicted on the generated future frames based on a 3D-CNN based saliency prediction model. 

In this paper, we propose a new hybrid model to predict gaze in egocentric videos, which combines bottom-up visual saliency with task-dependent attention transition. To the best of our knowledge, this is the first work to explore the patterns in attention transition for egocentric gaze prediction. 

\section{Gaze Prediction Model}

In this section, we first give overview of the network architecture of the proposed gaze prediction model, and then explain the details of each component. The details of training the model are provided in the end. 

\begin{figure}[t]
    \centering
    \includegraphics[width=\linewidth]{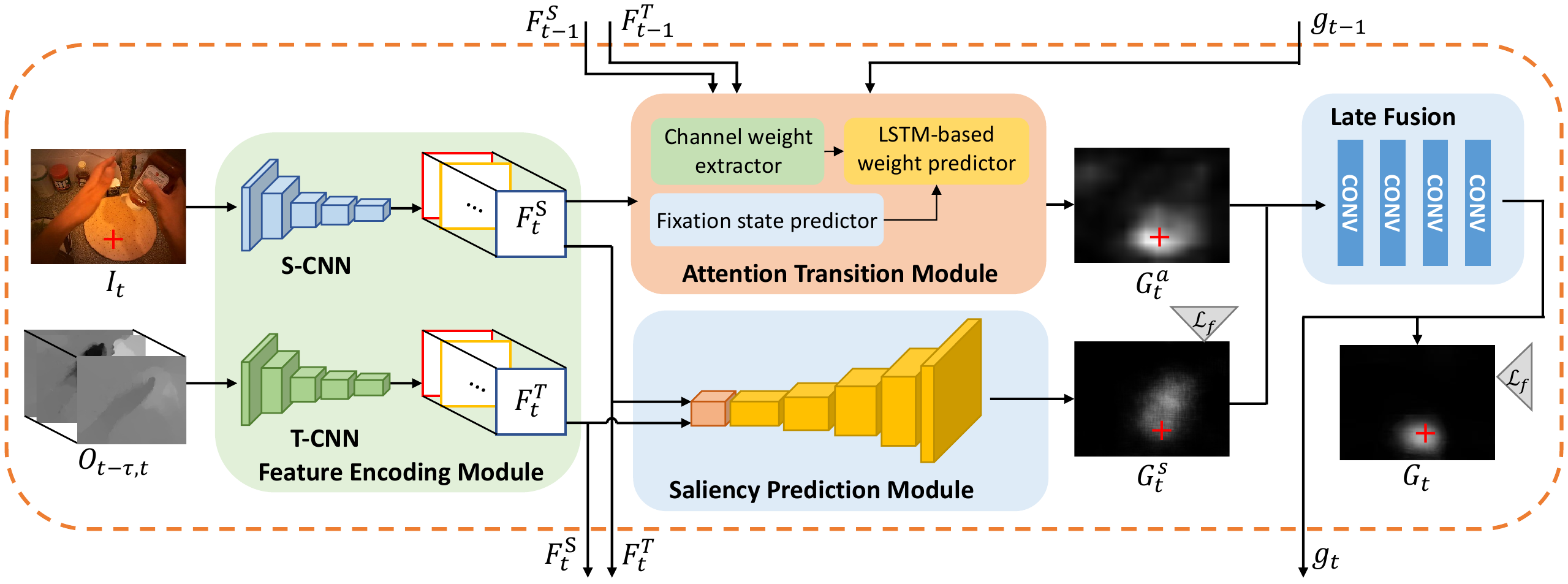}
    \caption{The architecture of our proposed gaze prediction model. The red crosses in the figure indicate ground truth gaze positions.}
    \label{fig:wholemodel}
\end{figure}

\subsection{Model Architecture}

Given consecutive video frames as input, we aim to predict a gaze position in each frame. To leverage both bottom-up visual saliency and task-dependent attention transition, we propose a hybrid model that 1) predicts a saliency map from each video frame, 2) predicts an attention map by exploiting temporal context of gaze fixations, and 3) fuses the saliency map and the attention map to output a final gaze map. 

The model architecture is shown in Figure \ref{fig:wholemodel}. The feature encoding module is composed by a spatial Convolutional Neural Network (S-CNN) and a temporal Convolutional Neural Network (T-CNN), which extract latent representations from a single RGB image and stacked optical flow images respectively. The saliency prediction module generates a saliency map based on the extracted latent representation. The attention transition module generates an attention map based on previous gaze fixations and head motion. The late fusion module combines the results of saliency prediction and attention transition to generate a final gaze map. The details of each module will be given in the following part.

\subsection{Feature Encoding}

At time $t$, the current video frame $I_t$ and stacked optical flow $O_{t-\tau, t}$ are fed into S-CNN and T-CNN to extract latent representations $F^{S}_t = h^S(I_t)$ from the current RGB frame, and $F^{T}_t = h^T(O_{t-\tau, t})$from the stacked optical flow images for later use. Here $\tau$ is fixed as 10 following \cite{simonyan2014two}.

The feature encoding network of S-CNN and T-CNN follows the base architecture of the first five convolutional blocks in Two Stream CNN \cite{simonyan2014two}, while omitting the final max pooling layer. We choose to use the output feature map of the last convolution layer from the 5-th convolutional group, i.e., \textit{conv5\_3}. Further analysis of different choices of deep feature maps from other layers is described in Section \ref{Experiment:AA}. 

\subsection{Saliency Prediction Module}

Biologically, human tends to gaze at an image region with high saliency, i.e., a region containing unique and distinctive visual features \cite{sugano2013appearance}. In the saliency prediction module of our gaze prediction model, we learn to generate a visual saliency map which reflects image regions that are likely to attract human gaze. We fuse the latent representations $F^{S}_t$ and $F^{T}_t$ as an input to a saliency prediction decoder (denoted as $S$) to obtain the initial gaze prediction map $G^s_t$ (Eq. \ref{equa:sp}). We use the ``3dconv + pooling" method of \cite{feichtenhofer2016convolutional} to fuse the two input feature streams. Since our task is different from \cite{feichtenhofer2016convolutional}, we modify the kernel sizes of the fusion part, which can be seen in detail in Section \ref{details}. The decoder outputs a visual saliency map with each pixel value within the range of $[0,1]$. Details of the architecture of the decoder is described in Section \ref{details}. The equation for generating the visual saliency map is:
\begin{equation}\label{equa:sp}
    G^s_t=S(F^S_t,F^T_t)
\end{equation}

However, a saliency map alone does not predict accurately where people actually look \cite{yamada2010can}, especially in egocentric videos of natural dynamic scenes where the knowledge of a task has a strong influence on human gaze. To achieve better gaze prediction, high-level knowledge about a task, such as which object is to be looked at and manipulated next, has to be considered.

\subsection{Attention Transition Module}

\begin{figure}[t]
    \centering
    \includegraphics[width=\linewidth]{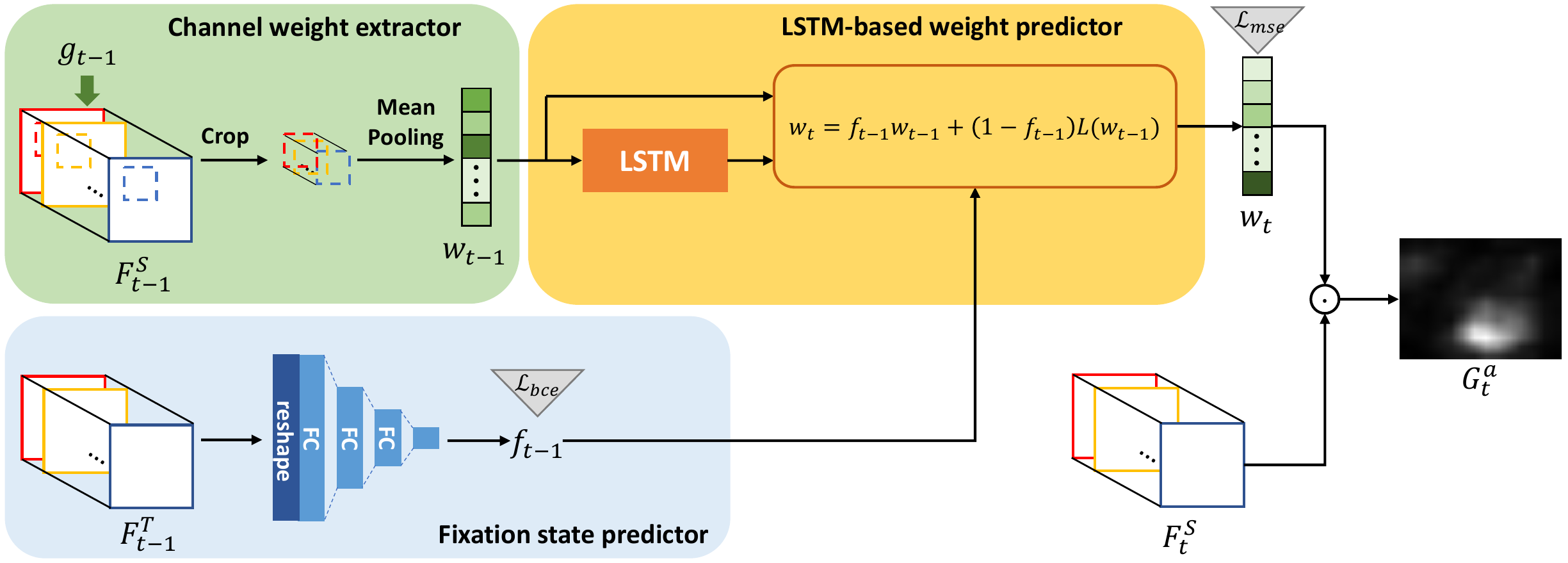}
    \caption{The architecture of the attention transition module.}
    \label{fig:lstm}
\end{figure}

During the procedure of performing a task, the task knowledge strongly influences the temporal transition of human gaze fixations on a series of objects. Therefore, given previous gaze fixations, it is possible to anticipate the image region where next attention occurs.
However, direct modeling the object transition explicitly such as using object categories is problematic since a reliable and generic object detector is needed.
Motivated by the fact that different channels of a feature map in top convolutional layers correspond well to spatial responses of different high-level semantics such as different object categories \cite{chen2016sca}\cite{zhou2016learning}, we represent the region that is likely to attract human attention by weighting each channel of the feature map differently. We train a Long Short Term Memory (LSTM) model \cite{lstm} to predict a vector of channel weights which is used to predict the region of attention at next fixation. 
Figure \ref{fig:lstm} depicts the framework of the proposed attention transition module. The module is composed of a channel weight extractor (C), a fixation state predictor (P), and a LSTM-based weight predictor (L). 

The channel weight extractor takes as input the latent representation $F^S_{t-1}$ and the predicted gaze point $g_{t-1}$ from the previous frame. $F^S_{t-1}$ is in fact a stack of feature maps with spatial resolution $14\times14$ and 512 channels. From each channel, we project the predicted gaze position $g_{t-1}$ onto the 14$\times$14 feature map, and crop a fixed size area with height $H_c$ and width $W_c$ centered at the projected gaze position. We then average the value of the cropped feature map at each channel, obtaining a 512-dimensional vector of channel weight $w_{t-1}$:
\begin{equation}
    w_{t-1} = C(F^S_{t-1}, g_{t-1})
\end{equation}
where $C(\cdot)$ indicates the cropping and averaging operation, $w_{t-1}$ is used as feature representation of the region of attention around the gaze point at frame $t-1$.

The fixation state predictor takes the latent representation of $F^T_{t-1}$ as input and outputs a probabilistic score of fixation state $f^p_{t-1}=P(F^T_{t-1}) \in [0,1]$. Basically, the score tells how likely fixation is occurring in the frame $t-1$. The fixation state predictor is composed by three fully connected layers followed by a final softmax layer to output a probabilistic score for gaze fixation state.

We use a LSTM to learn the attention transition by learning the transition of channel weights. The LSTM is trained based on a sequence of channel weight vectors extracted from images \textit{at the boundaries of} all gaze fixation periods with ground-truth gaze points, \textit{i.e.} we only extract one channel weight vector for each fixation to learn its transition between fixations. During testing, given a channel weight vector $w_{t-1}$, the trained LSTM outputs a channel weight vector $L(w_{t-1})$ that represents the region of attention at next gaze fixation. 
We also consider the dynamic behavior of gaze and its influence on attention transition. Intuitively speaking, during a period of fixation, the region of attention tends to remain unchanged, and the attended region changes only when saccade happens. Therefore, we compute the region of attention at current frame  $w_{t}$ as a linear combination of previous region of attention $w_{t-1}$ and the anticipated region of attention at next fixation $L(w_{t-1})$, weighted by the predicted fixation probability $f^p_{t-1}$: 
\begin{equation}
    w_t = f^p_{t-1}\cdot w_{t-1} + (1-f^p_{t-1})\cdot L(w_{t-1})
\end{equation}

Finally, an attention map $G^a_t$ is computed as the weighted sum of the latent representation $F^S_{t}$ at frame $t$ by using the resulting channel weight vector $w_t$:
\begin{equation}
    G^a_t= \sum_{c=1}^{n} w_t[c] \cdot F^S_{t}[c]
\end{equation}
where $[c]$ denotes the c-th dimension/channel of $w_t$/$F^S_{t}$ respectively.

\subsection{Late Fusion}

We build the late fusion module (LF) on top of the saliency prediction module and the attention transition module, which takes $G^s_t$ and $G^a_t$ as input and outputs the predicted gaze map $G_t$. 
\begin{equation}
    G_t = LF(G^s_t, G^a_t)
\end{equation}
Finally, a predicted 2D gaze position $g_t$ is given as the spatial coordinate of maximum value of $G_t$.

\subsection{Training}

For training gaze prediction in saliency prediction module and late fusion module, the ground truth gaze map $\hat{G}$ is given by convolving an isotropic Gaussian over the measured gaze position in the image. Previous work used either Binary Cross-Entropy loss \cite{kuen2016recurrent}, or KL divergence loss \cite{zhang2017deep} between the predicted gaze map and the ground truth gaze map for training neural networks. However, these loss functions do not work well with noisy gaze measurement. A measured gaze position is not static but continuously quivers in a small spatial range, even during fixation, and conventional loss functions are sensitive to small fluctuations of gaze. This observation motivates us to propose a new loss function, where the loss of pixels within small distance from the measured gaze position is down-weighted. More concretely, we modify the Binary Cross-Entropy loss function ($\mathcal{L}_{bce}$) across all the $N$ pixels with the weighting term $1+d_i$ as:
\begin{equation}\label{equa:2df}
    \mathcal{L}_{f}(G,\hat{G}) = -\frac{1}{N}\sum_{i=1}^{N}(1+d_i)\big\{\hat{G}[i]\cdot log(G[i]) + (1-\hat{G}[i])\cdot log(1-G[i])\big\}
\end{equation}
where $d_i$ is the euclidean distance between ground truth gaze position and the pixel $i$, normalized by the image width.

For training the fixation state predictor in the attention transition module, we treat the fixation prediction of each frame as a binary classification problem. Thus, we use the Binary Cross-Entropy loss function for training the fixation state predictor.
For training the LSTM-based weight predictor in the attention transition module, we use the mean squared error loss function across all the $n$ channels:
\begin{equation}
    \mathcal{L}_{mse} (w_t, \hat{w_t}) = \frac{1}{n}\sum_{i=1}^n(w_t[i] - \hat{w_t}[i])^2
\end{equation}
where $w_t[i]$ denotes the i-th element of $w_t$.

\subsection{Implementation details}\label{details}
We describe the network structure and training details in this section. Our implementation is based on the PyTorch~\cite{paszke2017automatic} library. The feature encoding module follows the base architecture of the first five convolutional blocks (\textit{conv1 $\sim$ conv5}) of VGG16~\cite{vgg} network. We remove the last max-pooling layer in the 5-th convolutional block. We initialize these convolutional layers using pre-trained weights on ImageNet \cite{deng2009imagenet}. Following \cite{simonyan2014two}, since the input channels of T-CNN is changed to 20, we average the weights of the first convolution layer of T-CNN part. The saliency prediction module is a set of 5 convolution layer groups following the inverse order of VGG16 while changing all max pooling layers into upsampling layers. We change the last layer to output 1 channel and add sigmoid activation on top. Since the input of the saliency prediction module contains latent representations from both S-CNN and T-CNN, we use a 3d convolution layer (with a kernel size of $1 \times 3 \times 3$) and a 3d pooling layer (with a kernel size of $2 \times 1 \times 1$) to fuse the inputs. Thus, the input and output sizes are all 224 $\times$ 224. The fixation state predictor is a set of fully connected (FC) layers, whose output sizes are 4096,1024,2 sequentially. The LSTM is a 3-layer LSTM whose input and output sizes are both 512. The late fusion module consists of 4 convolution layers followed by sigmoid activation. The first three layers have a kernel size of 3 $\times$ 3, 1 zero padding, and output channels 32,32,8 respectively, and the last convolution layer has a kernel size of 1 with a single output channel. We empirically set both the height $H_c$ and width $W_c$ for cropping the latent representations to be 3.

The whole model is trained using Adam optimizer \cite{kingma2014adam} with its default settings. We fix the learning rate as 1e-7 and first train the saliency prediction module for 5 epochs for the module to converge. We then fix the saliency prediction module and train the LSTM-based weight predictor and the fixation state predictor in the attention transition module. Learning rates for other modules in our framework are all fixed as 1e-4. After training the attention transition module, we fix the saliency prediction and the attention transition module to train the late fusion module in the end.

\section{Experiments}

We first evaluate our gaze prediction model on two public egocentric activity datasets (\textbf{GTEA Gaze} and \textbf{GTEA Gaze Plus}). We compare the proposed model with other state-of-the-art methods and provide detailed analysis of our model through ablation study and visualization of outputs of different modules. Furthermore, to examine our model's ability in learning attention transition, we visualize output of the attention transition module on a newly collected test set from GTEA Gaze Plus dataset (denoted as \textbf{GTEA-sub}).

\subsection{Datasets}
We introduce the datasets used for gaze prediction and attention transition.

\textbf{GTEA Gaze}
contains 17 video sequences of kitchen tasks performed by 14 subjects.  Each video clip lasts for about 4 minutes with the frame rate of 15 fps and an image resolution of 480 $\times$ 640. We use videos 1, 4, 6-22 as a training set and the rest as a test set as in Yin \emph{et al.} \cite{li2013learning}. 

\textbf{GTEA Gaze Plus}
contains 37 videos with the frame rate of 24 fps and an image resolution of 960 $\times$ 1280. In this dataset each of the 5 subjects performs 7 meal preparation activities in a more natural environment. Each video clip is 10 to 15 minute long on average. Similarly to \cite{li2013learning}, gaze prediction accuracy is evaluated with 5-fold cross validation across all 5 subjects.

\textbf{GTEA-sub}\label{gtea-sub}
contains 227 video frames selected from the sampled frames of GTEA Gaze Plus dataset. Each selected frame is not only under a gaze fixation, but also contains the object (or region) that is to be attended at the next fixation. We manually draw bounding boxes on those regions by inspecting future frames. The dataset is used to examine whether or not our model trained on GTEA Gaze Plus (excluding GTEA-sub) has successfully learned the task-dependent attention transition. 

\subsection{Evaluation Metrics}

We use two standard evaluation metrics for gaze prediction in egocentric videos: Area Under the Curve (AUC) \cite{AUC} and Average Angular Error (AAE) \cite{AAE}.
\textbf{AUC} is the area under a curve of true positive rate versus false positive rate for different thresholds on the predicted gaze map. It is a commonly used evaluation metric in saliency prediction. \textbf{AAE} is the average angular distance between the predicted and the ground truth gaze positions.

\subsection{Results on Gaze Prediction}

\subsubsection{Baselines.}
We use the following baselines for gaze prediction:
\begin{itemize}
    \item \textit{Saliency prediction algorithms}: We compare our method with several representative saliency prediction methods. More specifically, we used Itti's model \cite{itti2000saliency}, Graph Based Visual Saliency (GBVS \cite{harel2007graph}), and a deep neural network based saliency model as the current state of the art (SALICON \cite{huang2015salicon}). 
    \item \textit{Center bias}: Since egocentric gaze data is observed to have  a strong center bias, we use the image center as the predicted gaze position as in \cite{li2013learning}. 
    \item \textit{Gaze prediction algorithms}: We also compare our method with two state-of-the-art gaze prediction methods: the egocentric cue-based method (Yin \emph{et al.} \cite{li2013learning}), and the GAN-based method (DFG \cite{zhang2017deep}). Note that although the goal of \cite{zhang2017deep} is gaze anticipation in future frames, it also reported gaze prediction in the current frame.
\end{itemize}

\subsubsection{Performance Comparison.}

The quantitative results of different methods on two datasets are given in Table~\ref{table:gazepred}. Our method significantly outperforms all baselines on both datasets, particularly on the AAE score. Although there is only a small improvement on the AUC score, it can be seen that previous method of DFG \cite{zhang2017deep} has already achieved quite high score and the space of improvement is limited. Besides, we have observed from experiments that high AUC score does not necessarily mean high performance of gaze prediction.
The overall performance on GTEA Gaze is lower than that on GTEA Gaze Plus. The reason might be that the number of training samples in GTEA Gaze is smaller and over 25\% of ground truth gaze measurements are missing.
It is also interesting to see that the center bias outperforms all saliency-based methods and works only slightly worse than Yin \emph{et al.} \cite{li2013learning} on GTEA Gaze Plus, which demonstrates the strong spatial bias of gaze in egocentric videos.

\setlength{\tabcolsep}{4pt}
\begin{table}
\begin{center}
\caption{Performance comparison of different methods for gaze prediction on two public datasets. Higher AUC (or lower AAE) means higher performance.}
\label{table:gazepred}
\begin{tabular}{lcccc}
\toprule
\multirow{2}{*}{Metrics} & \multicolumn{2}{c}{GTEA Gaze Plus} & \multicolumn{2}{c}{GTEA Gaze} \\
\cmidrule(lr){2-3}\cmidrule(lr){4-5}
 & AAE (deg) & AUC & AAE (deg) & AUC \\
\midrule
Itti \emph{et al.} \cite{itti2000saliency} & 19.9 & 0.753 & 18.4 & 0.747 \\
GBVS \cite{harel2007graph} & 14.7 & 0.803 & 15.3 & 0.769 \\
SALICON \cite{huang2015salicon} & 15.6 & 0.818 & 16.5 & 0.761\\
Center bias & 8.6 & 0.819 & 10.2 & 0.789 \\
Yin \emph{et al.} \cite{li2013learning}  & 7.9 & 0.867 & 8.4 & 0.878\\
DFG \cite{zhang2017deep} & 6.6 & 0.952 & 10.5 & 0.883\\
Our full model & {\bf 4.0} & {\bf 0.957} & {\bf 7.6} & {\bf 0.898}
\\
\bottomrule
\end{tabular}
\end{center}
\end{table}
\setlength{\tabcolsep}{1.4pt}

\subsubsection{Ablation Study.}
To study the effect of each module of our model, and the effectiveness of our modified binary cross entropy loss (Equation \ref{equa:2df}), we conduct an ablation study and test each component on both GTEA Gaze Plus and GTEA Gaze datasets. Our baselines include: 1) single-stream saliency prediction with binary cross entropy loss (\textbf{S-CNN bce} and \textbf{T-CNN bce}), 2) single-stream saliency prediction with our modified bce loss (\textbf{S-CNN} and \textbf{T-CNN}), 3) two-stream saliency prediction with bce loss (\textbf{SP bce}), 4) two-stream input saliency prediction with our modified bce loss (\textbf{SP}), 5) the attention transition module (\textbf{AT}), and our full model. 

Table \ref{table:ablation} shows the results of the ablation study. The comparison of the same framework with different loss functions shows that our modified bce loss function is more suitable for the training of gaze prediction in egocentric video. The SP module performs better than either of the single-stream saliency prediction (S-CNN and T-CNN), indicating that both spatial and temporal information are needed for accurate gaze prediction. It is important to see that the AT module performs competitively or better than the SP module. This validates our claim that learning task-dependent attention transition is important in egocentric gaze prediction. More importantly, our full model outperforms all separate components by a large margin, which confirms that the bottom-up visual saliency and high-level task-dependent attention are complementary cues to each other and should be considered together in modeling human attention.

\setlength{\tabcolsep}{4pt}
\begin{table}
\begin{center}
\caption{Results of ablation study}
\label{table:ablation}
\begin{tabular}{lcccc}
\toprule
\multirow{2}{*}{Metrics} & \multicolumn{2}{c}{GTEA Gaze plus} & \multicolumn{2}{c}{GTEA Gaze} \\
\cmidrule(lr){2-3}\cmidrule(lr){4-5}
 & AAE (deg) & AUC & AAE (deg) & AUC \\
\midrule
S-CNN (bce) & 5.61 & 0.893 & 9.90 & 0.854 \\
T-CNN (bce) & 6.15 & 0.906 & 10.08 & 0.854\\
S-CNN  & 5.57 & 0.905 & 9.72 & 0.857 \\
T-CNN  & 6.07 & 0.906 & 9.6 & 0.859\\
SP (bce) & 5.63 & 0.918 & 9.53 & 0.860 \\
SP & 5.52 & 0.928 & 9.43 & 0.861\\
AT  & 5.02 & 0.940 & 9.51 & 0.857\\
Our full model & {\bf 4.05} & {\bf 0.957} & {\bf 7.58} & {\bf 0.898}
\\
\bottomrule
\end{tabular}
\end{center}
\end{table}
\setlength{\tabcolsep}{1.4pt}

\begin{figure}
    \centering
    \includegraphics[width=\linewidth]{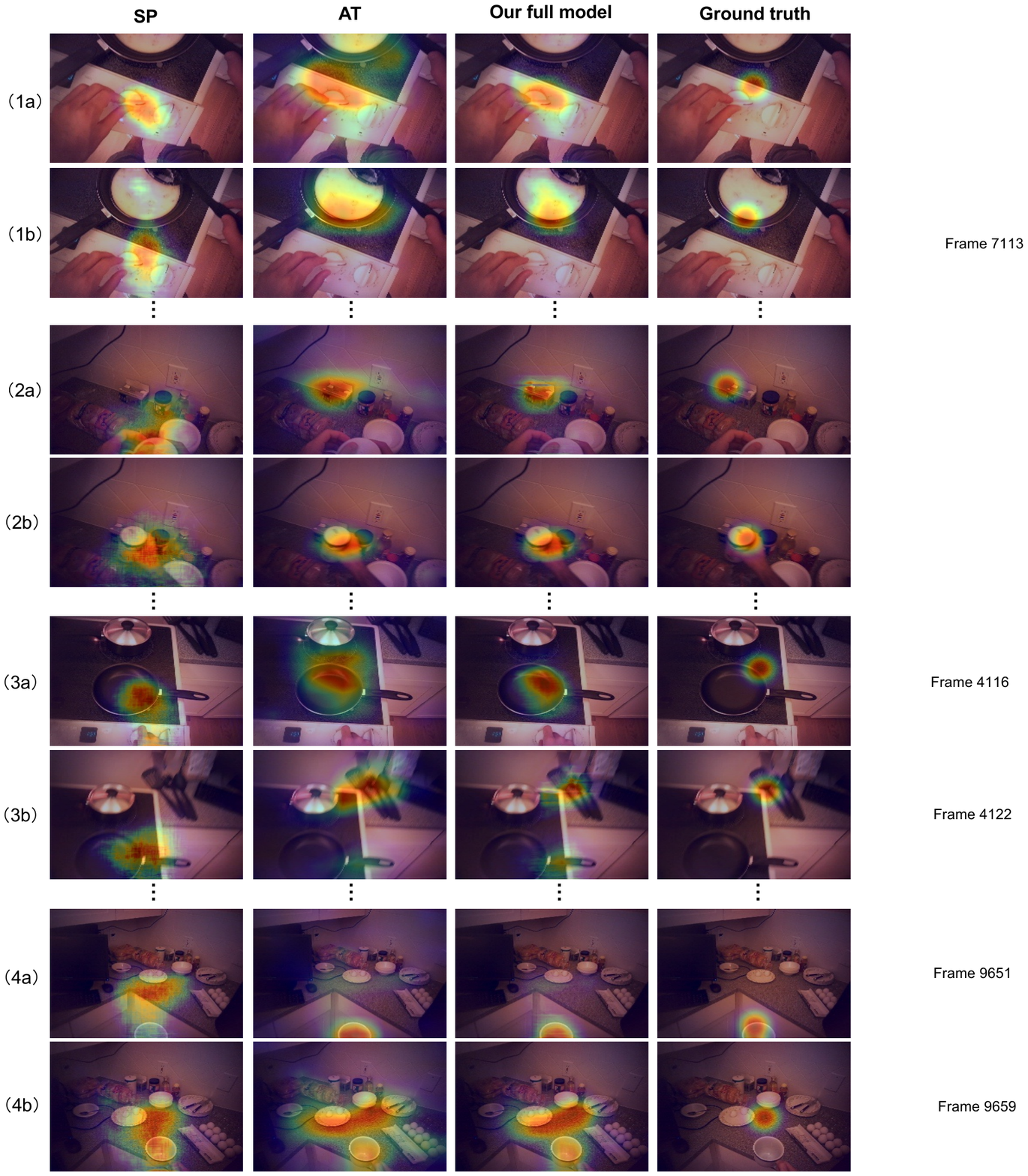}
    \caption{Visualization of predicted gaze maps from our model. Each group contains two images from two consecutive fixations, where a happens before b. We show the output heatmap from the saliency prediction module (\textbf{SP}) and the attention transition module (\textbf{AT}) as well as our full model. The ground truth gaze map (the rightmost column) is obtained by convolving an isotropic Gaussian on the measured gaze point.}
    \label{fig:gaze_quality}
\end{figure}

\subsubsection{Visualization.}
Figure \ref{fig:gaze_quality} shows qualitative results of our model. Group (1a, 1b) shows a typical gaze shift: the camera wearer shifts his attention to the pan after turning on the oven. SP fails to find the correct gaze position in (1b) only from visual features of the current frame. Since AT exploits the high-level temporal context of gaze fixations, it successfully predicts the region to be on the pan. Group (2a, 2b) demonstrates a ``put" action: the camera wearer first looks at the target location, then puts the can to that location. It is interesting that AT has learned the camera wearer's intention, and predicts the region at the target location rather than the more salient hand region in (2a). In group (3a, 3b), the camera wearer searches for a spatula after looking at the pan. Again, AT has learned this context which leads to more accurate gaze prediction than SP. Finally, group (4a, 4b) shows that SP and AT are complementary to each other. While AT performs better in (4a), and SP performs better in (4b), the full model combines the merits of both AT and SP to make better prediction. Overall, these results demonstrate that the attention transition plays an important role in improving gaze prediction accuracy.

\subsubsection{Cross Task Validation.}


To examine how the task-dependent attention transition learned in our model can generalize to different tasks under same (kitchen) scene, we perform a cross validation across the 7 different meal preparation tasks on GTEA Gaze Plus dataset. We consider the following experiment settings: 
\begin{itemize}
    \item \textbf{SP}: The saliency prediction module is treated as a generic component and trained on a separate subset of the dataset. We also use it as a baseline for studying the performance variation of different settings. 
    \item \textbf{AT\_d}: The attention transition module is trained and validated under different tasks. Average performance of 7-fold cross validation is reported.
    \item \textbf{AT\_s}: The attention transition module is trained and validated on two splits of the same task. Average performance of 7 tasks is reported.
    \item \textbf{SP+AT\_d}: The late fusion on top of \textbf{SP} and \textbf{AT\_d}.
    \item \textbf{SP+AT\_s}: The late fusion on top of \textbf{SP} and \textbf{AT\_s}.
\end{itemize}

\begin{figure}[t]
    \centering
    \includegraphics[width=0.9\linewidth]{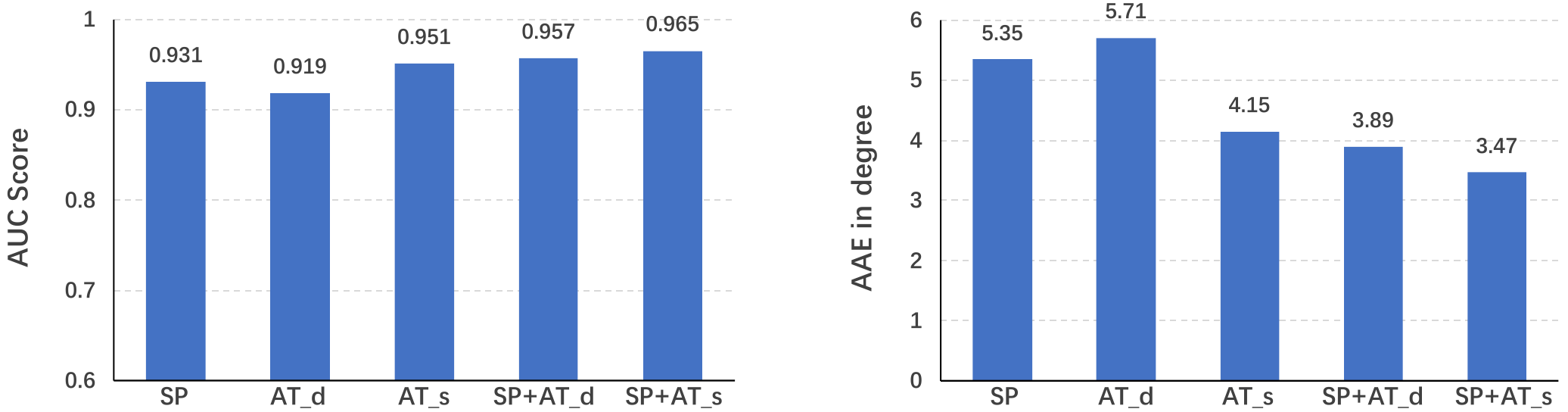}
    \caption{AUC and AAE scores of cross task validation. Five different experiment settings (explained in the text below) are compared to study the differences of attention transition in different tasks. 
    }
    \label{fig:crosstask}
\end{figure}

Quantitative results of different settings are shown in Figure \ref{fig:crosstask}. Both AUC and AAE scores show the same performance trend with different settings. AT\_d works worse than SP, while AT\_s outperforms SP. This is probably due to the differences of gaze behavior contained in different tasks. However, SP+AT\_d with the late fusion module can still improve the performance compared with SP and AT\_s, even with the context learned from different tasks. 

\subsection{Examination of the attention transition module}\label{Experiment:AA}


We further demonstrate that our attention transition module is able to learn meaningful transition between adjacent gaze fixations. This ability has important applications in computer-aided AR system, such as implying a person where to look next in performing a complex task. We conduct a new experiment on the GTEA-sub dataset (as introduced in Section \ref{gtea-sub}) to test the attention transition module of our model. 
Since here we focus on the module's ability of attention transition, we omit the fixation state predictor in the module and assume the output of the fixation state predictor as $f_{t}=0$ in the test frame. The module takes $w_{t}$ calculated from the region of current fixation as input and outputs an attention map on the same frame which represents the predicted region of the next fixation. We extract a 2D position from the maximum value of the predicted heatmap and calculate its rate of falling within the annotated bounding box as the transition accuracy. 

\begin{figure}[t]
    \centering
    \includegraphics[width=0.95\linewidth]{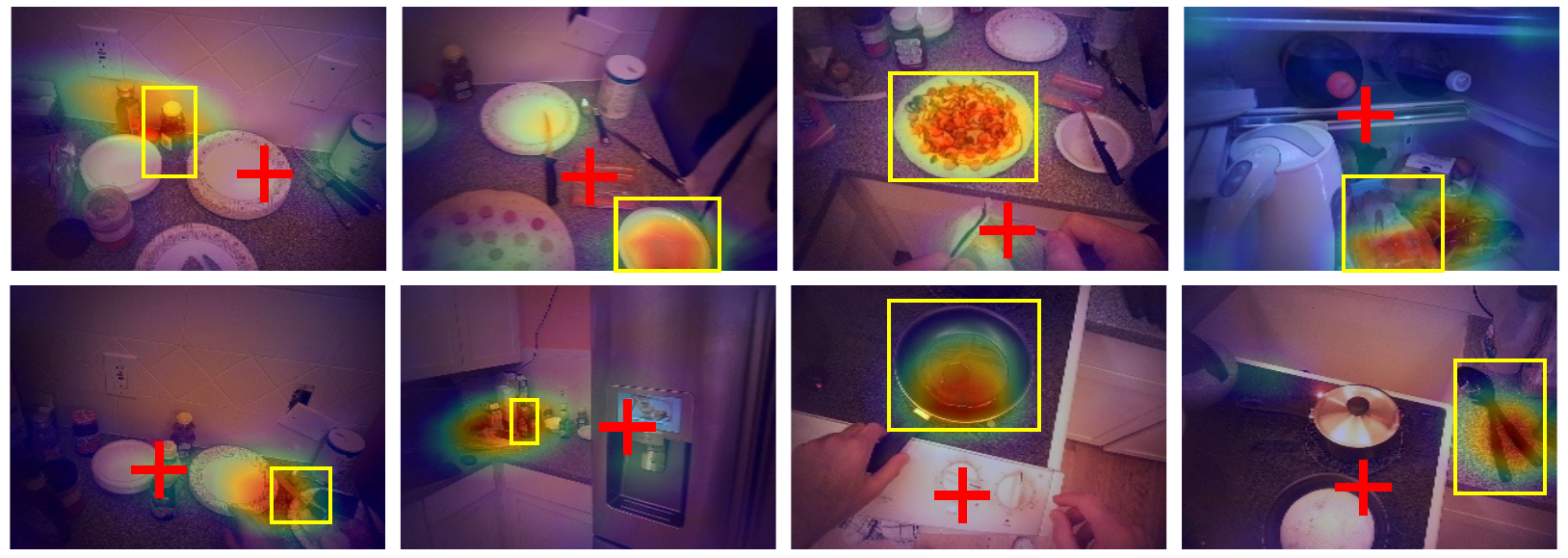}
    \caption{Qualitative results of attention transition. We visualize the predicted heatmap on the current frame, together with the current gaze position (red cross) and ground truth bounding box of the object/region of the next fixation (yellow box).}
    \label{fig:channel}
\end{figure}

We conduct experiments based on different latent representations extracted from the convolutional layer: \textit{conv5\_1}, \textit{conv5\_2}, and \textit{conv5\_3} of S-CNN. The accuracy based on the above three convolutional layers are 71.7\%, 83.0\%, and 86.8\% respectively, while the accuracy based on random position is 10.7\%. We also tried using random channel weight as the output of channel weight predictor to compute attention map based on the latent representation of \textit{conv5\_3}, and the accuracy is 9.4\%. This verifies that our model can learn meaningful attention transition of the performed task.
Figure \ref{fig:channel} shows some qualitative results of the attention transition module learned based on layer \textit{conv5\_3}. It can be seen that the attention transition module can successfully predict the image region of next fixation.

\section{Conclusion and Future Work}

This paper presents a hybrid model for gaze prediction in egocentric videos. Task-dependent attention transition is learned to predict human attention from previous fixations by exploiting the temporal context of gaze fixations. The task-dependent attention transition is further integrated with a CNN-based saliency model to leverage the cues from both bottom-up visual saliency and high-level attention transition. The proposed model achieves state-of-the-art performance in two public egocentric datasets. 

As for our future work, we plan to explore the task-dependent gaze behavior in a broader scale, \textit{i.e.} tasks in an office or in a manufacturing factory, and to study the generalizability of our model in different task domains.

\section*{Acknowledgments}
This work was supported by JST CREST Grant Number JPMJCR14E1, Japan.
%
%
%
\bibliographystyle{splncs04}
\bibliography{bib1321}

\begin{thebibliography}{10}
\providecommand{\url}[1]{\texttt{#1}}
\providecommand{\urlprefix}{URL }
\providecommand{\doi}[1]{https://doi.org/#1}

\bibitem{betancourt2015evolution}
Betancourt, A., Morerio, P., Regazzoni, C.S., Rauterberg, M.: The evolution of
  first person vision methods: A survey. IEEE Transactions on Circuits and
  Systems for Video Technology  \textbf{25}(5),  744--760 (2015)

\bibitem{borji2013state}
Borji, A., Itti, L.: State-of-the-art in visual attention modeling. IEEE
  transactions on pattern analysis and machine intelligence  \textbf{35}(1),
  185--207 (2013)

\bibitem{AUC}
Borji, A., Tavakoli, H.R., Sihite, D.N., Itti, L.: Analysis of scores,
  datasets, and models in visual saliency prediction. In: ICCV (2013)

\bibitem{cai2015scalable}
Cai, M., Kitani, K.M., Sato, Y.: A scalable approach for understanding the
  visual structures of hand grasps. In: ICRA (2015)

\bibitem{cai2016understanding}
Cai, M., Kitani, K.M., Sato, Y.: Understanding hand-object manipulation with
  grasp types and object attributes. In: Robotics: Science and Systems (2016)

\bibitem{cai2017ego}
Cai, M., Kitani, K.M., Sato, Y.: An ego-vision system for hand grasp analysis.
  IEEE Transactions on Human-Machine Systems  \textbf{47}(4),  524--535 (2017)

\bibitem{cai2018desktop}
Cai, M., Lu, F., Gao, Y.: Desktop action recognition from first-person
  point-of-view. IEEE Transactions on Cybernetics  (2018)

\bibitem{cao2015look}
Cao, C., Liu, X., Yang, Y., Yu, Y., Wang, J., Wang, Z., Huang, Y., Wang, L.,
  Huang, C., Xu, W., et~al.: Look and think twice: Capturing top-down visual
  attention with feedback convolutional neural networks. In: ICCV (2015)

\bibitem{chen2016sca}
Chen, L., Zhang, H., Xiao, J., Nie, L., Shao, J., Liu, W., Chua, T.S.: Sca-cnn:
  Spatial and channel-wise attention in convolutional networks for image
  captioning (2017)

\bibitem{deng2009imagenet}
Deng, J., Dong, W., Socher, R., Li, L.J., Li, K., Fei-Fei, L.: Imagenet: A
  large-scale hierarchical image database. In: CVPR (2009)

\bibitem{fathi2012learning}
Fathi, A., Li, Y., Rehg, J.M.: Learning to recognize daily actions using gaze.
  In: ECCV (2012)

\bibitem{feichtenhofer2016convolutional}
Feichtenhofer, C., Pinz, A., Zisserman, A.: Convolutional two-stream network
  fusion for video action recognition. In: CVPR (2016)

\bibitem{harel2007graph}
Harel, J., Koch, C., Perona, P.: Graph-based visual saliency. In: NIPS (2007)

\bibitem{lstm}
Hochreiter, S., Schmidhuber, J.: Long short-term memory. Neural computation
  \textbf{9}(8),  1735--1780 (1997)

\bibitem{hou2012image}
Hou, X., Harel, J., Koch, C.: Image signature: Highlighting sparse salient
  regions. IEEE transactions on pattern analysis and machine intelligence
  \textbf{34}(1),  194--201 (2012)

\bibitem{huang2015salicon}
Huang, X., Shen, C., Boix, X., Zhao, Q.: Salicon: Reducing the semantic gap in
  saliency prediction by adapting deep neural networks. In: ICCV (2015)

\bibitem{huang2017temporal}
Huang, Y., Cai, M., Kera, H., Yonetani, R., Higuchi, K., Sato, Y.: Temporal
  localization and spatial segmentation of joint attention in multiple
  first-person videos. In: ICCV Workshop (2017)

\bibitem{itti2000saliency}
Itti, L., Koch, C.: A saliency-based search mechanism for overt and covert
  shifts of visual attention. Vision research  \textbf{40}(10-12),  1489--1506
  (2000)

\bibitem{itti1998model}
Itti, L., Koch, C., Niebur, E.: A model of saliency-based visual attention for
  rapid scene analysis. IEEE Transactions on pattern analysis and machine
  intelligence  \textbf{20}(11),  1254--1259 (1998)

\bibitem{kingma2014adam}
Kingma, D.P., Ba, J.: Adam: A method for stochastic optimization. arXiv
  preprint arXiv:1412.6980  (2014)

\bibitem{kitani2011fast}
Kitani, K.M., Okabe, T., Sato, Y., Sugimoto, A.: Fast unsupervised ego-action
  learning for first-person sports videos. In: CVPR (2011)

\bibitem{kuen2016recurrent}
Kuen, J., Wang, Z., Wang, G.: Recurrent attentional networks for saliency
  detection. In: CVPR (2016)

\bibitem{land2004coordination}
Land, M.F.: The coordination of rotations of the eyes, head and trunk in
  saccadic turns produced in natural situations. Experimental brain research
  \textbf{159}(2),  151--160 (2004)

\bibitem{li2013learning}
Li, Y., Fathi, A., Rehg, J.M.: Learning to predict gaze in egocentric video.
  In: ICCV (2013)

\bibitem{lin2014saliency}
Lin, Y., Kong, S., Wang, D., Zhuang, Y.: Saliency detection within a deep
  convolutional architecture. In: AAAI Workshops (2014)

\bibitem{pan2016shallow}
Pan, J., Sayrol, E., Giro-i Nieto, X., McGuinness, K., O'Connor, N.E.: Shallow
  and deep convolutional networks for saliency prediction. In: CVPR (2016)

\bibitem{parkhurst2002modeling}
Parkhurst, D., Law, K., Niebur, E.: Modeling the role of salience in the
  allocation of overt visual attention. Vision research  \textbf{42}(1),
  107--123 (2002)

\bibitem{paszke2017automatic}
Paszke, A., Gross, S., Chintala, S., Chanan, G., Yang, E., DeVito, Z., Lin, Z.,
  Desmaison, A., Antiga, L., Lerer, A.: Automatic differentiation in pytorch
  (2017)

\bibitem{ramanishka2017top}
Ramanishka, V., Das, A., Zhang, J., Saenko, K.: Top-down visual saliency guided
  by captions. In: CVPR (2017)

\bibitem{AAE}
Riche, N., Duvinage, M., Mancas, M., Gosselin, B., Dutoit, T.: Saliency and
  human fixations: state-of-the-art and study of comparison metrics. In: ICCV
  (2013)

\bibitem{simonyan2013deep}
Simonyan, K., Vedaldi, A., Zisserman, A.: Deep inside convolutional networks:
  Visualising image classification models and saliency maps. arXiv preprint
  arXiv:1312.6034  (2013)

\bibitem{simonyan2014two}
Simonyan, K., Zisserman, A.: Two-stream convolutional networks for action
  recognition in videos. In: NIPS (2014)

\bibitem{vgg}
Simonyan, K., Zisserman, A.: Very deep convolutional networks for large-scale
  image recognition. arXiv preprint arXiv:1409.1556  (2014)

\bibitem{sugano2013appearance}
Sugano, Y., Matsushita, Y., Sato, Y.: Appearance-based gaze estimation using
  visual saliency. IEEE transactions on pattern analysis and machine
  intelligence  \textbf{35}(2),  329--341 (2013)

\bibitem{treisman1980feature}
Treisman, A.M., Gelade, G.: A feature-integration theory of attention.
  Cognitive psychology  \textbf{12}(1),  97--136 (1980)

\bibitem{xu2015gaze}
Xu, J., Mukherjee, L., Li, Y., Warner, J., Rehg, J.M., Singh, V.: Gaze-enabled
  egocentric video summarization via constrained submodular maximization. In:
  CVPR (2015)

\bibitem{yamada2010can}
Yamada, K., Sugano, Y., Okabe, T., Sato, Y., Sugimoto, A., Hiraki, K.: Can
  saliency map models predict human egocentric visual attention? In: ACCV
  (2010)

\bibitem{yamada2011attention}
Yamada, K., Sugano, Y., Okabe, T., Sato, Y., Sugimoto, A., Hiraki, K.:
  Attention prediction in egocentric video using motion and visual saliency.
  In: Pacific-Rim Symposium on Image and Video Technology. pp. 277--288.
  Springer (2011)

\bibitem{zhang2017deep}
Zhang, M., Teck~Ma, K., Hwee~Lim, J., Zhao, Q., Feng, J.: Deep future gaze:
  Gaze anticipation on egocentric videos using adversarial networks. In: CVPR
  (2017)

\bibitem{zhao2015saliency}
Zhao, R., Ouyang, W., Li, H., Wang, X.: Saliency detection by multi-context
  deep learning. In: CVPR (2015)

\bibitem{zhou2016learning}
Zhou, B., Khosla, A., Lapedriza, A., Oliva, A., Torralba, A.: Learning deep
  features for discriminative localization. In: CVPR (2016)

\end{thebibliography}

\end{document}